\documentclass[letterpaper]{article} 
\usepackage{aaai2026}  
\usepackage{times}  
\usepackage{helvet}  
\usepackage{courier}  
\usepackage[hyphens]{url}  
\usepackage{graphicx} 
\usepackage{natbib}  
\usepackage{caption} 
\usepackage{algorithm}
\usepackage{algorithmic}
\usepackage{multirow}
\usepackage {tcolorbox}
\usepackage{booktabs}

\usepackage{newfloat}
\usepackage{listings}
\DeclareCaptionStyle{ruled}{labelfont=normalfont,labelsep=colon,strut=off} 
\floatstyle{ruled}
\newfloat{listing}{tb}{lst}{}
\floatname{listing}{Listing}
\title{PRIME: Planning and Retrieval-Integrated Memory for Enhanced Reasoning}
\author{Hieu Tran$^{1,2}$, Zonghai Yao$^{1,2}$, Nguyen Tran$^2$, Zhichao Yang $^2$,  Feiyun Ouyang$^{1,3}$, Shuo Han$^3$,  Razieh Rahimi$^2$, Hong Yu$^{1,2,3}$\\
$^{1}$Center for Healthcare Organization and Implementation Research, VA Bedford Health Care  \\
$^2$ Manning College of Information and Computer Sciences, University of Massachusetts Amherst\\
$^3$ Miner School of Computer and Information Sciences, University of Massachusetts Lowell\\
}

\usepackage{bibentry}

\begin{document}

\maketitle

\begin{abstract}
Inspired by the dual-process theory of human cognition from \textit{Thinking, Fast and Slow}, we introduce \textbf{PRIME} (Planning and Retrieval-Integrated Memory for Enhanced Reasoning), a multi-agent reasoning framework that dynamically integrates \textbf{System 1} (fast, intuitive thinking) 
and \textbf{System 2} (slow, deliberate thinking). 
PRIME first employs a Quick Thinking Agent (System 1)
to generate a rapid answer; if uncertainty is detected, it then triggers a structured System 2 reasoning pipeline composed of specialized agents for \textit{planning}, \textit{hypothesis generation}, \textit{retrieval}, \textit{information integration}, and \textit{decision-making}. This multi-agent design faithfully mimics human cognitive processes and enhances both efficiency and accuracy. Experimental results with LLaMA 3 models demonstrate that PRIME enables open-source LLMs to perform competitively with state-of-the-art closed-source models like GPT-4 and GPT-4o on benchmarks requiring multi-hop and knowledge-grounded reasoning. This research establishes PRIME as a scalable solution for improving LLMs in domains requiring complex, knowledge-intensive reasoning.
\end{abstract}


\begin{figure*}
    \centering
    \includegraphics[width=0.9\linewidth]{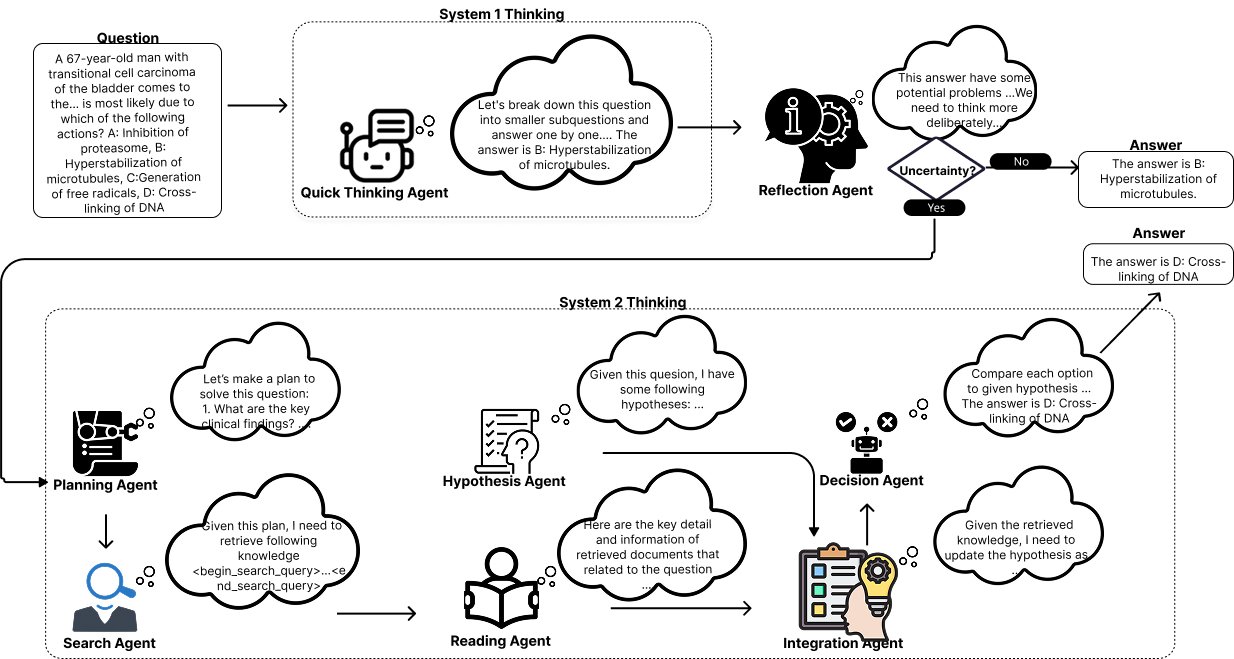}
    \caption{Overview of our reasoning process. The framework mimics human dual-system cognition by integrating fast, intuitive reasoning (System 1) and slower, deliberative reasoning (System 2).}
    \label{fig:overview}
\end{figure*}

\section{Introduction}

Human cognition is marked by its remarkable versatility, effortlessly transitioning from instant intuitive judgments such as recognizing faces or interpreting emotional cues to deliberate analytical reasoning involved in solving complex puzzles or making detailed plans. According to the influential dual-process theory popularized by Daniel Kahneman in \textit{Thinking, Fast and Slow} \cite{kahneman2011thinking}, these cognitive abilities stem from two fundamentally different modes of thought: \textit{System 1}, which operates quickly, intuitively, and with minimal cognitive effort; and \textit{System 2}, which is slower, more deliberate, analytical, and capable of managing demanding cognitive tasks \cite{evans1984heuristic, kahneman2003maps}.

System 1 continuously generates intuitions, impressions, and rapid judgments, often sufficient for routine or straightforward decisions. However, complex tasks, those involving multi-step planning, retrieval of external knowledge, and careful evidence integration necessitate engaging the more effortful System 2~\cite{huang2022towards, qiao2022reasoning, wang2022towards, shaikh2022second}. However, even these complex reasoning tasks are rarely executed entirely by System 2 alone; instead, most human reasoning processes fluidly integrate rapid intuitive responses with occasional deeper reflection, triggered when intuition proves insufficient or potentially erroneous~\cite{evans1984heuristic, kahneman2003maps, kahneman2011thinking}.

Motivated by this understanding of human cognitive efficiency, we introduce \textbf{PRIME} (Planning and Retrieval-Integrated Memory for Enhanced Reasoning), a multi-agent reasoning framework designed explicitly to mimic this dual-process model. PRIME initially employs a \textit{Quick Thinking Agent} (System 1) to produce intuitive answers rapidly. Subsequently, a specialized \textit{Reflection Agent} performs explicit self-reflection to critically evaluate the intuitive response. If self-reflection identifies uncertainty, potential errors, or risk of hallucinations, PRIME activates a more deliberative reasoning process (System 2), orchestrating specialized agents responsible for systematic planning, hypothesis generation, targeted knowledge retrieval, information integration, and comprehensive reasoning. This adaptive mechanism ensures efficient use of computational resources, engaging slower, computationally intensive reasoning only when necessary, significantly reducing unnecessary deliberation.

This design also addresses a crucial limitation observed in standalone intuitive or deliberative approaches. While fast intuitive reasoning (System 1) is computationally efficient, it often suffers from inaccuracies or hallucinations. Conversely, detailed analytical reasoning (System 2), although more powerful, is resource-intensive and occasionally susceptible to overly elaborate or misleading reasoning paths. By effectively integrating both cognitive modes through targeted self-reflection, PRIME capitalizes on their respective strengths, enhancing overall accuracy while mitigating the inherent shortcomings of each mode when used individually. To validate the effectiveness and versatility of PRIME, we conducted extensive experiments in several challenging reasoning benchmarks, including MedQA, MedMCQA, MMLU-Medical, Musique, 2Wiki, HotpotQA, and Amboss. Experimental results clearly demonstrate that PRIME enables open-source large language models, such as LLaMA 3, to achieve performance competitive with advanced closed-source counterparts, such as GPT-4 and GPT-4o. Detailed analyses further highlight that PRIME efficiently allocates computational resources, strategically triggering System 2 reasoning predominantly for more challenging tasks, while confidently relying on Zrapid System 1 responses for simpler questions.

In summary, our work makes the following key contributions:
\begin{enumerate}
    \item Based on our review of prior dual-system reasoning architectures and cognitive-inspired frameworks, we are the first to propose a multi-agent reasoning framework that explicitly operationalizes the dual-process theory of cognition by structuring agents into fast (System 1) and slow (System 2) reasoning processes, mediated through explicit self-reflection.
    \item We demonstrate that selective activation of deliberative reasoning significantly enhances computational efficiency, improving performance by avoiding unnecessary deliberation and reducing hallucinations inherent to purely intuitive reasoning.
    \item Empirical evaluations across multiple challenging benchmarks confirm PRIME’s superior reasoning accuracy and computational efficiency, establishing it as a robust and scalable framework for complex, knowledge-intensive reasoning tasks.
\end{enumerate}

\section{Related Work}

\noindent\textbf{Multi-agent reasoning frameworks} have been widely explored to enhance the robustness and compositionality of language model-based systems. Recent work frames LLMs as agents that collaborate \citep{yao2023react, yang2023auto}, debate \citep{du2023improving, menick2022teaching}, or act hierarchically through planning and execution \citep{liu2023agentbench, liu2023llm+, zhou2023webarena}. For example, ReAct \citep{yao2023react} combines reasoning and acting by allowing agents to interleave thoughts and tool use, while Self-Refine \citep{madaan2303self} introduces feedback loops for iterative self-improvement. AutoGPT and related tool-augmented frameworks \citep{schick2023toolformer, shinn2023reflexion} orchestrate multiple agents for autonomous task completion. PRIME is inspired by this agentic structure but differs in its cognitive framing: agents in PRIME are hierarchically organized into fast and slow thinkers, rather than peers or collaborators, and only slow agents are triggered when fast thinking is insufficient.

\noindent\textbf{Dual-system cognitive models} trace back to the "System 1 vs. System 2" framework popularized by Kahneman \citep{kahneman2011thinking}, and have motivated AI systems that balance rapid heuristics with deliberate reasoning. Recent LLM research reflects this divide: Chain-of-Thought prompting \citep{wei2022chain} and Tree-of-Thoughts \citep{yao2023tree} promote structured multi-step reasoning akin to System 2, while strategies like self-consistency \citep{wang2022self}, scratchpad reasoning \citep{nye2021show}, and iterative refinement \citep{madaan2303self} aim to improve correctness through slow, revisable computation. Cognitive architectures such as meta-reasoning controllers \citep{hong2023metagpt} have also modeled reasoning as a two-phase process. PRIME unifies these views by combining fast, subquestion-driven intuitive answers with a deliberative reasoning system (planning, retrieval, hypothesis testing) that is selectively triggered via reflection, enabling both efficient and robust inference.

\noindent\textbf{System 1 and System 2 hybrid reasoning} has also emerged as a guiding paradigm in recent LLM work. The influential survey by \citet{li2025system} categorizes reasoning LLMs into architectural and training strategies that emulate System 2 cognition: structured exploration (e.g., MCTS in Marco-o1~\citep{zhao2024marco}), macro actions (e.g., ReasonFlux~\citep{yang2025reasonflux}), and self-improvement via feedback loops (e.g., rStar-Math~\citep{guan2025rstar}). While these models implicitly follow dual-system principles, they lack an explicit \textit{separation} of intuitive vs. deliberative agents. In contrast, PRIME operationalizes this distinction directly: System 1 conducts rapid subquestion-based reasoning; System 2, composed of modular slow agents, engages only when reflection detects flaws or uncertainty.

\section{Methodology}

Our framework is explicitly inspired by the dual-process cognitive theory introduced in \textit{Thinking, Fast and Slow}. PRIME closely mirrors this cognitive process: when presented with a question, the framework first rapidly generates an intuitive answer through the \textit{Quick Thinking Agent} (System 1). Then, the \textit{Reflection Agent} critically evaluates this intuitive response, by explicitly performing self-reflection to determine whether the intuitive answer is reliable or if it potentially contains errors, logical inconsistencies, or uncertainties. If the Reflection Agent recognizes potential problems or uncertainties, it triggers the deeper and deliberate reasoning process (System 2) as shown in Figure \ref{fig:overview}.


\begin{figure}[!ht]
    \centering
    \includegraphics[width=1.0\linewidth]{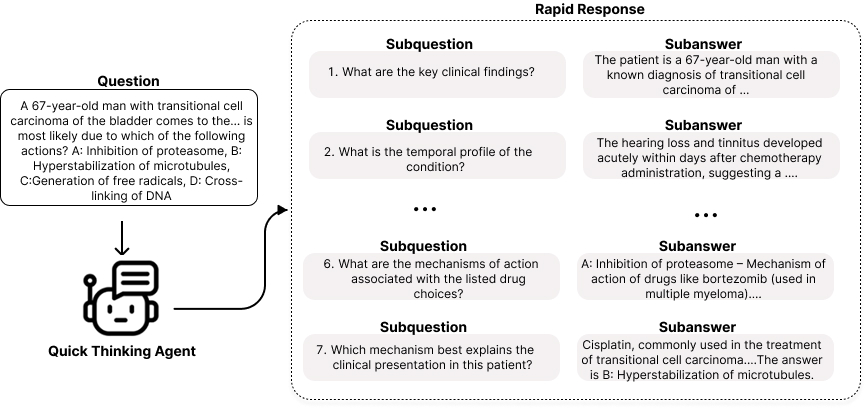}
    \caption{Quick Thinking Agent for System 1 reasoning. Upon receiving a question, the Quick Thinking Agent rapidly decomposes it into a series of subquestions and answers each one sequentially. }
    \label{fig:quick}
\end{figure}

Once activated, PRIME’s System 2 involves structured reasoning steps, including explicit planning to break down the problem, targeted search and reading processes to mimic human memory recall from external knowledge sources, as well as structured hypothesis formulation and testing. Finally, a comprehensive decision-making step synthesizes information from memory and hypotheses to produce a robust, reliable final answer. In the following subsections, we detail each individual component of PRIME’s multi-agent architecture, highlighting their specific roles, interactions, and contributions to the overall reasoning process.


\subsection{System 1: Quick Intuitive Thinking}



The \textbf{Quick Thinking Agent} implements the fast, intuitive reasoning pathway inspired by \textbf{System 1} in human cognition. Unlike conventional Chain-of-Thought (CoT) prompting, which proceeds step-by-step in a linear generative fashion, our approach takes a more structured route. Specifically, the Quick Thinking Agent generates a structured response that \textit{decomposes the question into a sequence of subquestions and corresponding subanswers}, all within a single output. Each subquestion targets a specific facet of the problem, and its subanswer is generated immediately before progressing to the next. This structure enables efficient, context-aware reasoning while maintaining coherence across substeps. (Figure \ref{fig:quick}). The process continues iteratively, with the final subquestion aiming to synthesize all previous information and directly address the original question. This structured subquestion-based reasoning allows for rapid response generation while encouraging a minimal form of internal reflection across subcomponents. Beyond efficiency, this structured format also enhances transparency and verifiability. By explicitly surfacing intermediate reasoning steps, the subquestion–subanswer layout makes it easier for the subsequent Reflection Agent to detect potential flaws. It can examine the reasoning trace at a finer granularity, identify inconsistencies, and determine whether deeper deliberation is needed. While the Quick Thinking Agent is effective for many cases, it may still produce hallucinated or overconfident conclusions. These risks are mitigated through self-reflection and conditional triggering of System 2, discussed in the next section.
\begin{figure}[!ht]
    \centering
    \includegraphics[width=1.0\linewidth]{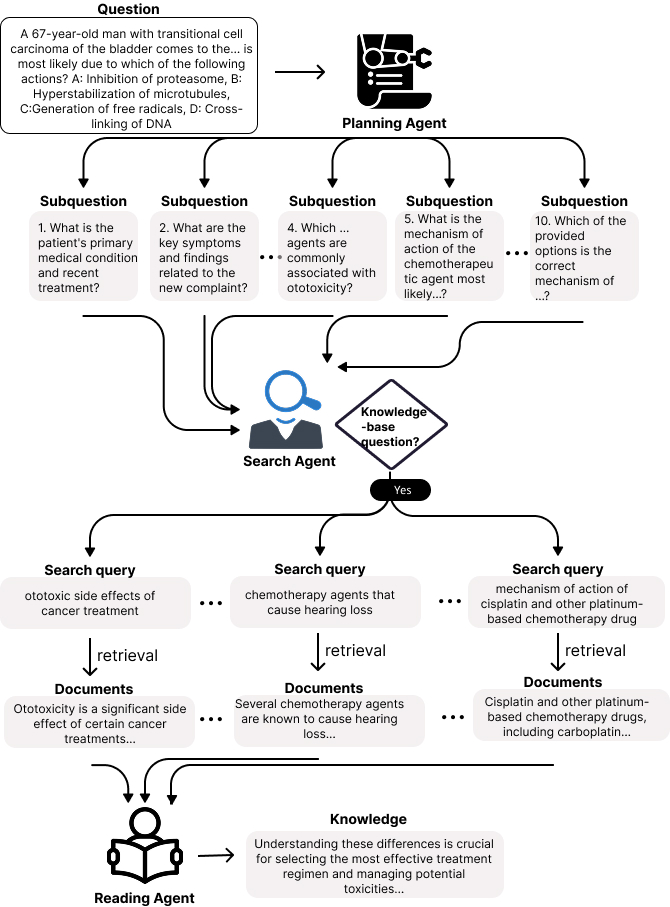}
    \caption{Memory recall process in System 2 reasoning. When System 2 is triggered, the Planning Agent decomposes the question into targeted subquestions. For each subquestion requiring external knowledge, the Search Agent issues domain-specific queries and retrieves relevant documents. The Reading Agent then distills key information from the retrieved evidence. This mimics human memory recall, where reasoning is guided by selectively retrieving and integrating key facts rather than exact memorization of full documents.}
    \label{fig:plan}
\end{figure}

\subsection{Reflection Agent: Self-Evaluation and Triggering Deliberation}

After the Quick Thinking Agent produces a complete answer through structured subquestioning, the \textbf{Reflection Agent} engages in a critical self-assessment of the generated response. This process draws inspiration from human metacognition—our ability to reflect on our own thinking, detect errors, and determine when deeper reasoning is warranted. Unlike approaches that rely solely on model confidence or heuristic scoring, the Reflection Agent performs explicit \textit{self-reflection} by analyzing the internal structure and content of the System 1 output. In particular, the subquestion–subanswer format generated by the Quick Thinking Agent offers a fine-grained reasoning trace, enabling the Reflection Agent to assess each step's logical coherence, consistency, and evidentiary support. It evaluates whether key information was ignored, whether each subanswer logically follows from its subquestion. If the Reflection Agent determines that the answer is well-supported and internally consistent, the system returns the System 1 output directly, completing the reasoning process efficiently. However, if it detects uncertainty, inconsistencies, hallucinations, or unsupported conclusions, it \textit{triggers System 2 reasoning}. This adaptive mechanism ensures that PRIME invokes computationally expensive deliberation only when necessary, balancing performance and efficiency while enhancing overall robustness.

\subsection{System 2: Memory Recall — Planning, Search, and Reading Agents}

When the Reflection Agent determines that deeper reasoning is needed, PRIME activates its System 2 pipeline, beginning with the \textbf{Memory Recall stage}. This stage is implemented through the coordinated actions of three specialized agents: the \textit{Planning Agent}, the \textit{Search Agent}, and the \textit{Reading Agent}. Together, they emulate the human process of recalling information—not by retrieving verbatim content from memory, but by reconstructing key facts, relevant concepts, and critical details from external sources (Figure \ref{fig:plan}).

In human cognition, memory recall is rarely a literal reproduction of textbook content. Instead, it involves abstracting important takeaways and reconstructing useful information relevant to the problem at hand. PRIME mirrors this process: it does not depend on cached memory or pre-learned patterns alone, but dynamically reads and reconstructs evidence from an external corpus, functioning like a human drawing upon both personal knowledge and external references.

\begin{figure}[!ht]
    \centering
    \includegraphics[width=1.0\linewidth]{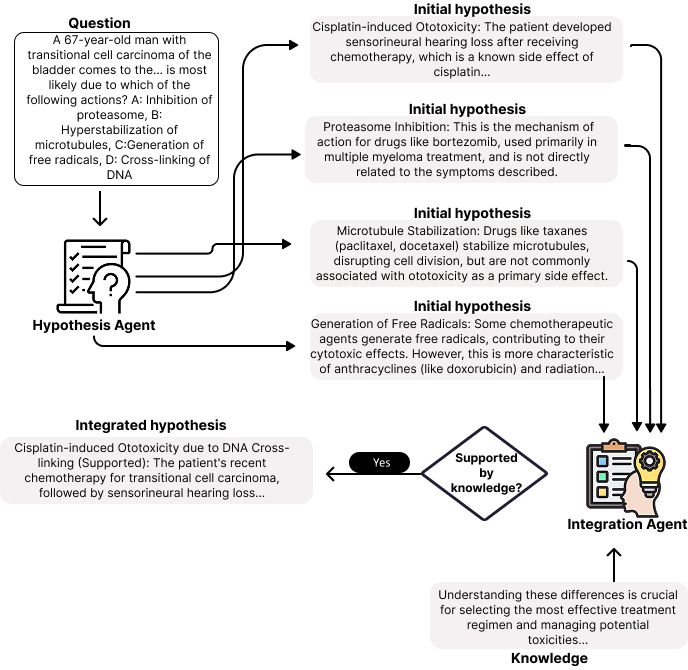}
    \caption{Hypothesis generation and testing in System 2 reasoning. The Hypothesis Agent formulates multiple initial hypotheses based on the question. The Integration Agent then evaluates each hypothesis by aligning it with key evidence from the memory recall phase. This process mimics human scientific reasoning by generating, testing, and selecting the most plausible explanation.}
    \label{fig:hypothesis}
\end{figure}

\noindent\textbf{Planning Agent:}
The Planning Agent initiates this process by decomposing the original question into a series of fine-grained subquestions. Each subquestion is designed to probe a specific piece of knowledge necessary to solve the original task. These subquestions guide the overall reasoning strategy and identify what kind of information needs to be retrieved.

\noindent\textbf{Search Agent:}  
The Search Agent formulates explicit search queries and retrieves relevant documents from an external knowledge base (e.g., medical textbooks, scientific articles) for subquestions requiring factual or knowledge-intensive answers. It determines whether a subquestion is answerable using internal knowledge or if external retrieval is necessary, mimicking the human ability to decide when to "look something up."

\noindent\textbf{Reading Agent:}  
Once relevant documents are retrieved, the Reading Agent processes and interprets their content. Rather than memorizing everything, the Reading Agent extracts and summarizes the key insights, facts, or clinical principles relevant to the reasoning task. This abstraction of retrieved knowledge mirrors how humans condense information into usable insights for problem-solving.

Together, these agents enable PRIME to simulate human-like memory recall by reading, filtering, and organizing knowledge around the task at hand. This design not only increases factual reliability but also reduces hallucinations by grounding responses in verifiable external evidence.

\subsection{System 2: Hypothesis Testing — Hypothesis and Integration Agents}

After the memory recall stage provides relevant knowledge from external sources, PRIME proceeds to the second phase of System 2 thinking: \textbf{Hypothesis Testing}. This stage mimics the human reasoning strategy of generating potential explanations and systematically testing them against known facts and observations. In PRIME, this functionality is handled by two agents: the \textit{Hypothesis Agent} and the \textit{Integration Agent} (Figure \ref{fig:hypothesis}).

\noindent\textbf{Hypothesis Agent:}  
Given the original question, the Hypothesis Agent formulates a set of \textbf{initial hypotheses} corresponding to each of the answer choices or plausible explanations. These hypotheses are concise interpretations or assumptions that could explain the observed situation. Each hypothesis attempts to bridge the gap between the clinical presentation and a potential mechanism of action, diagnosis, or treatment rationale. This stage reflects the human habit of mentally proposing candidate answers before investigating them further.

\noindent\textbf{Integration Agent:}  
Once the hypotheses are proposed, the Integration Agent takes over to test their validity. It does so by cross-referencing each hypothesis with the key knowledge retrieved during the memory recall stage. This agent examines whether each hypothesis is \textit{supported}, \textit{refuted}, or \textit{left inconclusive} based on the retrieved evidence. In doing so, it mimics a human's analytical process of weighing each candidate's answer against known facts or principles. If sufficient supporting evidence exists, the Integration Agent marks the hypothesis as valid and incorporates it into an \textbf{integrated hypothesis}, a refined conclusion that combines the initial assumption with supporting knowledge. This final hypothesis is used downstream in the decision stage to justify the selected answer. This two-agent setup enables PRIME to reason in a human-like manner by generating and testing ideas rather than relying on static rules or memorized answers. It also serves as a critical checkpoint in minimizing hallucinations, as each candidate explanation must be substantiated by external evidence before it can inform the final answer.

\subsection{System 2: Decision-Making — Decision Agent}

The final stage of System 2 reasoning in PRIME is handled by the \textbf{Decision Agent}, which is responsible for synthesizing all previous reasoning steps and selecting the most likely answer to the original question. After the Integration Agent produces one or more \textit{integrated hypotheses}, each of which combines a candidate explanation with supporting evidence, the Decision Agent evaluates these hypotheses in the context of the original question. The Decision Agent ranks the candidate hypotheses and selects the best-supported answer. This stage mimics the final human decision-making process: after reflecting on possible explanations and validating them against memory and evidence, we choose the most convincing answer based on logical fit and justification. This final decision is passed back as the output of the System 2 reasoning process, completing the reflective pipeline.

\section{Experiments}
In this section~\footnote{Detail settings of the experiments, descriptions of the evaluation tasks and baselines can be found in Appendix.}, we evaluate the performance of our proposed method on both medical reasoning and commonsense reasoning tasks using two large language models: LLaMA 3.1 8B Instruct and LLaMA 3.3 70B Instruct \cite{dubey2024LLaMA}, which we drop the word "Instruct" onward for simplicity. 

\begin{table}[ht]
\centering
\scalebox{0.58}{
\begin{tabular}{l|c|c|c|c}
\textbf{Model} & \textbf{MedQA} & \textbf{MedMCQA} & \textbf{MMLU-Medical} & \textbf{avg.} \\ \hline
\textbf{LLaMA3.1 8B} & & & & \\
\quad \quad CoT           & 61.51 & 55.15 & 71.63 & 62.76 \\
\quad \quad SC            & 64.73 & 56.35 & 72.73 & 64.60 \\
\quad \quad MedRAG        & 63.00 & 56.87 & 74.56 & 64.81 \\
\quad \quad i-MedRAG      & 73.61 & 61.61 & 78.42 & 71.21 \\
\quad \quad Search-O1     & 73.13 & 62.13 & 79.16 & 71.47 \\
\quad \quad PRIME         & \textbf{76.91} & \textbf{67.49} & \textbf{83.56} & \textbf{75.99} \\ \hline
\textbf{LLaMA3.3 70B} & & & & \\
\quad \quad CoT           & 76.51 & 68.28 & 81.36 & 75.38 \\
\quad \quad SC            & 79.73 & 70.69 & 82.37 & 77.59 \\
\quad \quad MedRAG        & 80.36 & 71.38 & 84.66 & 78.80 \\
\quad \quad i-MedRAG      & 81.82 & 72.54 & 86.69 & 80.35 \\
\quad \quad Search-O1     & 83.17 & 73.11 & 87.23 & 81.17 \\
\quad \quad PRIME         & \textbf{87.51} & \textbf{78.94} & \textbf{92.74} & \textbf{86.39} \\ \hline
Meditron 70B              & 51.69 & 46.74 & 64.92 & 54.45 \\ \hline
Mixtral (8x7B)            & 64.12 & 56.28 & 74.01 & 64.80 \\ \hline
GPT-3.5                   & 65.04 & 55.25 & 72.91 & 64.40 \\ \hline
GPT-4                     & 83.97 & 69.88 & 89.44 & 81.10 \\ \hline
GPT-4o-mini               & 73.29 & 66.17 & 84.31 & 74.59 \\ \hline
GPT-4o                    & 85.55 & 74.71 & 90.45 & 83.57 \\ 
\end{tabular}
}
\caption{Performance of PRIME and baseline methods on three medical reasoning benchmarks: MedQA, MedMCQA, and MMLU-Medical. SC denotes self-consistency decoding.}
\label{tab:medical_reasoning_comparison}
\end{table}

\subsection{Performance on Medical Reasoning Tasks}

Table~\ref{tab:medical_reasoning_comparison} presents the performance of PRIME and state-of-the-art baseline models on three challenging medical reasoning benchmarks: MedQA, MedMCQA, and MMLU-Medical. These benchmarks require not only complex multi-hop reasoning but also a high degree of factual precision, making them ideal for evaluating PRIME’s dual-system reasoning capabilities. The results demonstrate the effectiveness of PRIME in enhancing the reasoning abilities of LLaMA models compared to baseline methods, including Chain-of-Thought (CoT), Self-Consistency (SC), MedRAG, i-MedRAG, and Search-O1. Across both LLaMA3.1 8B and LLaMA3.3 70B, PRIME consistently outperforms all baselines. 

On the LLaMA3.1 8B model, PRIME achieves an average score of 75.99, surpassing the state-of-the-art i-MedRAG (71.21) and Search-O1 (71.47) by over 4 points (absolute), indicating the benefits of structured reasoning over purely retrieval-augmented approaches. On the larger LLaMA3.3 70B model, PRIME achieves an average of 86.39, setting a new state-of-the-art among open-source models. Notably, PRIME-enhanced LLaMA3.3 70B outperforms GPT-4 (81.10) and GPT-4o (83.57) on MedQA and MMLU-Medical, demonstrating its competitive edge even against top-tier closed-source models.


\subsection{Performance on Multi-hop Reasoning}

\begin{table}[ht]
\centering
\scalebox{0.65}{
\begin{tabular}{l|cc|cc|cc}
\textbf{Model} & \multicolumn{2}{c|}{\textbf{Musique}} & \multicolumn{2}{c|}{\textbf{2Wiki}} & \multicolumn{2}{c}{\textbf{HotpotQA}} \\
              & \textbf{EM} & \textbf{F1} & \textbf{EM} & \textbf{F1} & \textbf{EM} & \textbf{F1} \\ \hline
\textbf{LLaMA3.3 70B} & & & & & & \\
\quad \quad Naive RAG     & 19.14 & 30.52 & 33.64 & 38.22 & 30.33 & 40.06 \\
\quad \quad IRCoT         & 24.27 & 33.69 & 44.43 & 51.53 & 37.86 & 42.28 \\
\quad \quad Iter-RetGen   & 27.49 & 36.11 & 47.59 & 54.22 & 34.36 & 44.22 \\
\quad \quad RAG Agent     & 28.97 & 40.41 & 60.74 & 72.34 & 39.81 & 51.28 \\
\quad \quad Search-O1     & 30.37 & 41.94 & 63.33 & 74.24 & 41.68 & 54.81 \\
\quad \quad PRIME        & \textbf{35.17} & \textbf{48.81} & \textbf{68.84} & \textbf{79.81} & \textbf{46.51} & \textbf{60.68} \\ \hline
gpt-4o-mini               & 26.27 & 35.69 & 45.65 & 52.37 & 40.86 & 53.27 \\
gpt-4o                   & 33.91 & 47.38 & 64.51 & 73.56 & 45.94 & 56.67 \\
\end{tabular}
}
\caption{Exact Match (EM) and F1 scores on multi-hop QA benchmarks: Musique, 2Wiki, and HotpotQA.}
\label{tab:multihop_qa_results}
\end{table}

Table~\ref{tab:multihop_qa_results} reports the performance of PRIME and baseline methods on three multi-hop reasoning benchmarks: Musique, 2Wiki, and HotpotQA. Across all datasets, PRIME consistently outperforms baseline methods, including retrieval-augmented approaches such as RAG Agent, Search-O1, and Iter-RetGen. On Musique, PRIME achieves an EM score of 35.17 and an F1 score of 48.81, improving significantly over Search-O1 (30.37 EM, 41.94 F1). On 2Wiki, PRIME achieves 68.84 EM and 79.81 F1, surpassing strong baselines like RAG Agent and Search-O1 by a large margin. Similarly, on HotpotQA, PRIME achieves 46.51 EM and 60.68 F1, outperforming Search-O1 and approaching the performance of GPT-4o.



\begin{table}[h!]
\centering
\scalebox{0.6}{
\begin{tabular}{l|c}
\textbf{Configuration} & \textbf{Accuracy} \\ \hline
PRIME (System 1 + System 2) & \textbf{87.2} \\
System 1 & 80.4 \\
System 2 (Full) & 86.0 \\
System 2 (Planning + Search + Hypothesis + Integration + Decision) & 84.8 \\
System 2 (Planning + Search + Reading + Hypothesis + Decision) &	84.2 \\
System 2 (Planning + Search + Hypothesis + Decision) & 82.8 \\
System 2 (Planning + Search + Reading + Decision) 	& 83.6 \\
System 2 (Planning + Search + Decision) & 81.4 \\
System 2 (Hypothesis + Decision) & 80.6 \\
\end{tabular}}
\caption{Ablation study on PRIME components, evaluated on 250 MedQA samples using LLaMA 3.3 70B}
\label{tab:ablation_study}
\end{table}

\subsection{Ablation Study}

Table~\ref{tab:ablation_study} presents an ablation study analyzing the contribution of different components in PRIME, evaluated on 250 MedQA samples using LLaMA 3.3 70B. The full PRIME framework, integrating both System 1 and System 2, achieves the best performance at 87.2\%, demonstrating the advantage of combining rapid intuition with selective deep reasoning. Using only System 1 achieves 80.4\%, indicating that while quick subquestion decomposition is effective for easier cases, it struggles without deeper validation. Full System 2 reasoning alone reaches 86.0\%, confirming the strength of deliberate multi-step reasoning, although slightly less efficient than the combined PRIME system.

We further analyze various partial System 2 configurations. The "Planning + Search + Hypothesis + Integration + Decision" setting, which removes the Reading Agent, achieves 84.8\%, showing that document summarization helps but is not critical. Removing the Integration Agent ("Planning + Search + Reading + Hypothesis + Decision") drops accuracy to 84.2\%, suggesting hypothesis refinement plays an important role. Skipping both reading and integration ("Planning + Search + Hypothesis + Decision") reduces performance further to 82.8\%. Without hypothesis generation ("Planning + Search + Reading + Decision"), accuracy falls to 83.6\%, and removing both hypothesis generation and reading ("Planning + Search + Decision") results in 81.4\%. Finally, directly moving from hypothesis to decision without planning or retrieval ("Hypothesis + Decision") yields 80.6\%, confirming that external knowledge grounding is crucial.


\begin{table}[ht]
\centering
\scalebox{0.6}{
\begin{tabular}{l|c|c|c|c}
\textbf{Difficulty Level} & \textbf{Answering System} & \textbf{Correct} & \textbf{Incorrect} & \textbf{Accuracy (\%)} \\
\hline
\multirow{2}{*}{Very Easy} 
    & System 1 & 85 & 3 & 96.59 \\
    & System 2 & 12 & 0 & 100.00 \\
\hline
\multirow{2}{*}{Easy} 
    & System 1 & 76 & 6 & 92.68 \\
    & System 2 & 14 & 4 & 77.78 \\
\hline
\multirow{2}{*}{Medium} 
    & System 1 & 60 & 10 & 85.71 \\
    & System 2 & 24 & 6 & 80.00 \\
\hline
\multirow{2}{*}{Hard} 
    & System 1 & 28 & 22 & 56.00 \\
    & System 2 & 32 & 18 & 64.00 \\
\hline
\multirow{2}{*}{Very Hard} 
    & System 1 & 20 & 36 & 35.71 \\
    & System 2 & 24 & 20 & 54.55 \\
\end{tabular}
} 
\caption{Performance of System 1 and System 2 on Amboss across five difficulty levels. System 2 is triggered more frequently on harder questions, while System 1 performs well on easier questions.}
\label{tab:amboss_results}
\end{table}

\subsection{When System 2 Thinking Is Triggered?}

We conduct an analysis to understand when System 2 thinking is triggered by evaluating PRIME on 100 questions from each difficulty level (Very Easy, Easy, Medium, Hard, Very Hard) sampled from the Amboss question bank. Amboss is a comprehensive resource widely used by medical students and professionals, providing an extensive array of Step 1, Step 2 CK, and Step 3-style clinical questions, making it an ideal benchmark for evaluating reasoning under varying levels of complexity.

As shown in Table~\ref{tab:amboss_results}, System 1 is highly effective for easier questions. On Very Easy and Easy levels, System 1 achieves 96.59\% and 92.68\% accuracy, respectively, and the Reflection Agent rarely triggers System 2, since the intuitive answers are sufficient. When System 2 is occasionally triggered on easy questions, it slightly underperforms System 1 (77.78\% accuracy), suggesting that deep reasoning is often unnecessary for straightforward tasks. However, as question difficulty increases, System 2 is triggered more frequently. For Medium, Hard, and Very Hard questions, System 1 accuracy drops significantly, from 85.71\% down to 35.71\%, while System 2 helps recover performance, achieving 80.00\%, 64.00\%, and 54.55\% respectively. This shows that PRIME’s Reflection Agent correctly identifies when intuitive thinking is insufficient and selectively invokes deeper reasoning processes involving planning, retrieval, and hypothesis testing.

Overall, this analysis highlights that PRIME effectively mirrors human cognitive strategies: fast, efficient intuition is used when sufficient, while System 2 slow thinking is selectively triggered to handle more complex, knowledge-intensive problems, improving robustness without unnecessary computational overhead.

\begin{figure}[!ht]
    \centering
    \includegraphics[width=1.0\linewidth]{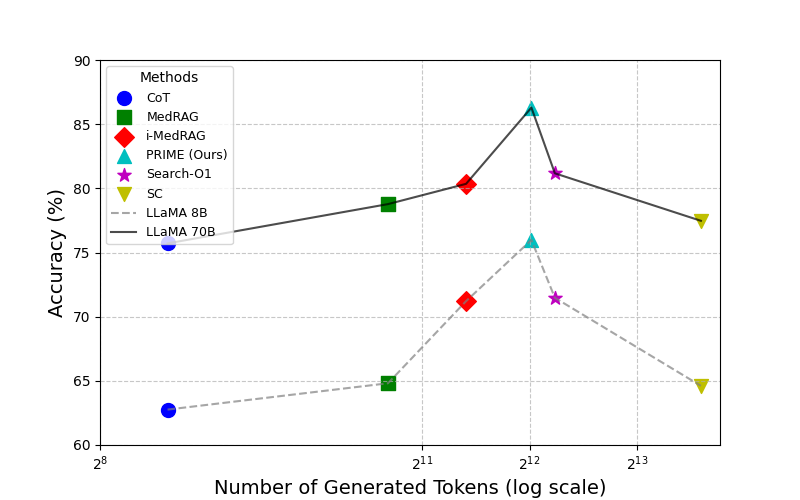}
    \caption{Accuracy vs. Number of Generated Tokens for Different Methods}
    \label{fig:comp}
\end{figure}

\subsection{Computational Analysis}

To assess the efficiency of different reasoning strategies, we analyze the trade-off between accuracy and the number of generated tokens, a proxy for computational cost. Figure~\ref{fig:comp} plots the average accuracy across three medical reasoning tasks against the number of generated tokens (log scale) for several methods under both LLaMA 8B and LLaMA 70B backbones. As shown, PRIME achieves the best accuracy on LLaMA 70B while maintaining a relatively modest token budget compared to other multi-step approaches. While Self-Consistency (SC) and Search-O1 also perform well, they incur significantly higher computational cost due to repeated sampling. In contrast, PRIME leverages System 1 to answer easier questions quickly and triggers System 2 reasoning only when necessary, resulting in a more efficient use of tokens. On LLaMA 8B, PRIME continues to outperform MedRAG and CoT by a wide margin in accuracy, while remaining competitive in token usage. i-MedRAG, despite sharing similar retrieval components, lacks structured planning and hypothesis testing, leading to lower accuracy despite similar token usage. These results highlight that PRIME’s dual-system design enables accurate reasoning with selective computational expenditure. 
By combining fast responses with targeted deep reasoning, PRIME offers a practical and scalable solution, particularly beneficial in deployment settings where compute resources are limited.

\begin{table}[ht]
\centering
\scalebox{0.99}{
\begin{tabular}{l|c}
\textbf{Agent} & \textbf{Score (\%)} \\ \hline
Reflection Agent    & 91 \\
Planning   Agent    & 96 \\
Search     Agent    & 94 \\
Reading    Agent    & 93 \\
Hypothesis    Agent & 82 \\
Integration    Agent & 84 \\
Decision       Agent & 87 \\
\end{tabular}
}
\caption{Expert evaluation accuracy of individual System 2 agents over 100 MedQA questions.}
\label{tab:expert_eval}
\end{table}

\subsection{Human Evaluation of System 2 Thinking}

To better understand the quality and reliability of individual agents within the System 2 pipeline, we conducted an expert evaluation using outputs from 100 MedQA questions. Two medical experts who had passed the USMLE were asked to independently assess each agent's contribution in terms of factual accuracy, appropriateness of reasoning, and medical soundness.  Each agent's output was rated as correct or incorrect for its respective task, and agreement was resolved through discussion. The results, shown in Table~\ref{tab:expert_eval}, indicate that upstream components—particularly Planning, Search, and Reading—exhibit high reliability (above 93\%), while Hypothesis and Integration agents were more error-prone, likely due to the inherent challenge of forming and testing hypotheses in ambiguous clinical contexts. Notably, the Reflection agent showed strong performance in identifying when slow reasoning is needed.

\section{Conclusion}

We presented \textbf{PRIME}, a dual-system, multi-agent reasoning framework that mimics human cognition by combining fast, intuitive responses with selectively triggered slow, deliberative reasoning. PRIME uses a lightweight System 1 for efficient subquestion decomposition and a reflective mechanism to invoke System 2 only when deeper reasoning is needed—engaging planning, retrieval, and hypothesis testing. Experiments on medical and multi-hop QA benchmarks show that PRIME significantly improves both accuracy and efficiency, enabling open-source LLMs to perform competitively with top closed-source models like GPT-4o. Ablation and difficulty-wise analyses further validate the value of dynamic reasoning depth and structured agent collaboration. PRIME offers a scalable and cognitively grounded approach to LLM reasoning, bridging fast heuristics and deliberate computation to support robust, adaptive decision-making.

\section{Limitations}

While PRIME demonstrates strong performance and efficiency across various reasoning benchmarks, it is not without limitations. First, the success of PRIME relies heavily on the effectiveness of its Reflection Agent to detect uncertainty in System 1 responses. If the agent fails to trigger System 2 when necessary, critical errors may go uncorrected.

Second, although System 2 introduces structured planning and evidence-based hypothesis testing, our error analysis reveals that the model is often anchored to its initial hypotheses. This anchoring limits its flexibility in updating beliefs even when contradictory evidence is presented, suggesting a need for more robust belief revision mechanisms.

Third, while effective, the modular multi-agent setup introduces additional latency and complexity compared to single-pass reasoning approaches. Although the system is designed to invoke computationally intensive reasoning selectively, optimizing the interaction and scheduling between agents remains an open challenge.

Finally, PRIME has been evaluated primarily in QA-style reasoning tasks. Its generalizability to broader task formats—such as open-ended generation, summarization, or scientific synthesis—requires further investigation and adaptation.

\section*{Acknowledgments}

This material is the result of work supported with resources and the use of facilities at the Center for Healthcare Organization and Implementation Research, VA Bedford Health Care.





\bibliography{aaai2026}



\end{document}